\documentclass[journal]{IEEEtran}
\usepackage{amsmath,amsfonts}
\usepackage{algorithmic}
\usepackage{algorithm}
\usepackage{array}
\usepackage{hyperref}
\usepackage[table,xcdraw]{xcolor}
\usepackage[caption=false,font=scriptsize,labelfont=rm,textfont=rm]{subfig}
\usepackage{textcomp}
\usepackage{pdfpages}
\usepackage{stfloats}
\usepackage{url}
\usepackage{float}
\usepackage{pifont}
\usepackage{longtable}
\usepackage{verbatim}
\usepackage{graphicx}
\usepackage{fancyhdr}
 \usepackage[switch]{lineno}
\usepackage{subfig}
\usepackage{indentfirst}
\usepackage{cite}

\usepackage{booktabs}
\usepackage{makecell}
\usepackage[numbers,sort&compress]{natbib}

\usepackage{multirow}
\usepackage{bbding}

\hypersetup{
    colorlinks=true,
    linkcolor=red,
    filecolor=black,
    urlcolor=black,
    }

\begin{document}

\title{Transformer RGBT Tracking with Spatio-Temporal Multimodal Tokens}

\author{Dengdi Sun, Yajie Pan, Andong Lu, Chenglong Li*, Bin Luo, \emph{Senior Member, IEEE}
\thanks{This work was supported in part by the National Natural Science Foundation of China (No. 62076005, U20B2068, 61906002); the Natural Science Foundation of Anhui Province (2008085MF191, 2008085QF306); and the University Synergy Innovation Program of Anhui Province, China (GXXT-2021-002).  (\emph{*Corresponding author is Chenglong Li}) }
\thanks{D. Sun and C. Li are with the Key Laboratory of Intelligent Computing \& Signal Processing, Ministry of Education, School of Artificial Intelligence, Anhui University, Hefei 230601, China, and Institute of Artificial Intelligence, Hefei Comprehensive National Science Center, Hefei 230026, China, Email: sundengdi@163.com, lcl1314@foxmail.com.}
\thanks{Y. Pan, A. Lu and B. Luo are with the Anhui Provincial Key Laboratory of Multimodal Cognitive Computation, School of Computer Science and Technology, Anhui University, Hefei, China, 230601, Email: pyj2897022134@163.com, adlu\_ah@foxmail.com, luobin@ahu.edu.cn.}
}

\markboth{Journal of \LaTeX\ Class Files,~Vol.~14, No.~8, August~2021}
{Shell \MakeLowercase{\textit{et al.}}: A Sample Article Using IEEEtran.cls for IEEE Journals}
\maketitle
\thispagestyle{fancy}
\lhead{}
\lfoot{}
\cfoot{}
\rfoot{}
\begin{abstract}
Many RGBT tracking researches primarily focus on modal fusion design, while overlooking the effective handling of target appearance changes. While some approaches have introduced historical frames or fuse and replace initial templates to incorporate temporal information, they have the risk of disrupting the original target appearance and accumulating errors over time. To alleviate these limitations, we propose a novel Transformer RGBT tracking approach, which mixes spatio-temporal multimodal tokens from the static multimodal templates and multimodal search regions in Transformer to handle target appearance changes, for robust RGBT tracking. We introduce independent dynamic template tokens to interact with the search region, embedding temporal information to address appearance changes, while also retaining the involvement of the initial static template tokens in the joint feature extraction process to ensure the preservation of the original reliable target appearance information that prevent deviations from the target appearance caused by traditional temporal updates. We also use attention mechanisms to enhance the target features of multimodal template tokens by incorporating supplementary modal cues, and make the multimodal search region tokens interact with multimodal dynamic template tokens via attention mechanisms, which facilitates the conveyance of multimodal-enhanced target change information. Our module is inserted into the transformer backbone network and inherits joint feature extraction, search-template matching, and cross-modal interaction. Extensive experiments on three RGBT benchmark datasets show that the proposed approach maintains competitive performance compared to other state-of-the-art tracking algorithms while running at 39.1 FPS. 
\end{abstract}

\section{Introduction}
\IEEEPARstart{V}{isual} object tracking is a fundamental but challenging research topic in computer vision. It facilitates numerous practical applications such as visual surveillance, driverless technology, and so on. 
Compared to single-modal tracking~\cite{weighted,wu2017rgb,xu2017learning,li2017weighted}, RGBT tracking harnesses the complementary information between modalities. 
The visible (RGB) modal provides rich details such as color and texture but is limited by lighting conditions and cannot capture clear images at night.
In contrast, the thermal infrared (TIR) modal is not affected by lighting and weather conditions, allowing it to operate in complete darkness, but it has a lower resolution and lacks color and texture details.
The integration of RGB and TIR modal cues enables visual trackers to achieve accurate and robust performance in challenging scenarios, such as illumination variation, background clutter, and bad weather. 
Hence, the RGBT tracking has garnered significant attention from researchers.
Moreover, large-scale RGBT tracking benchmark datasets~\cite{li2021lasher,li2019rgb}  have recently emerged, providing a better platform for the evaluation and development of RGBT trackers.
\begin{figure}[t]
    \centering
    \includegraphics[width=1.0\linewidth]{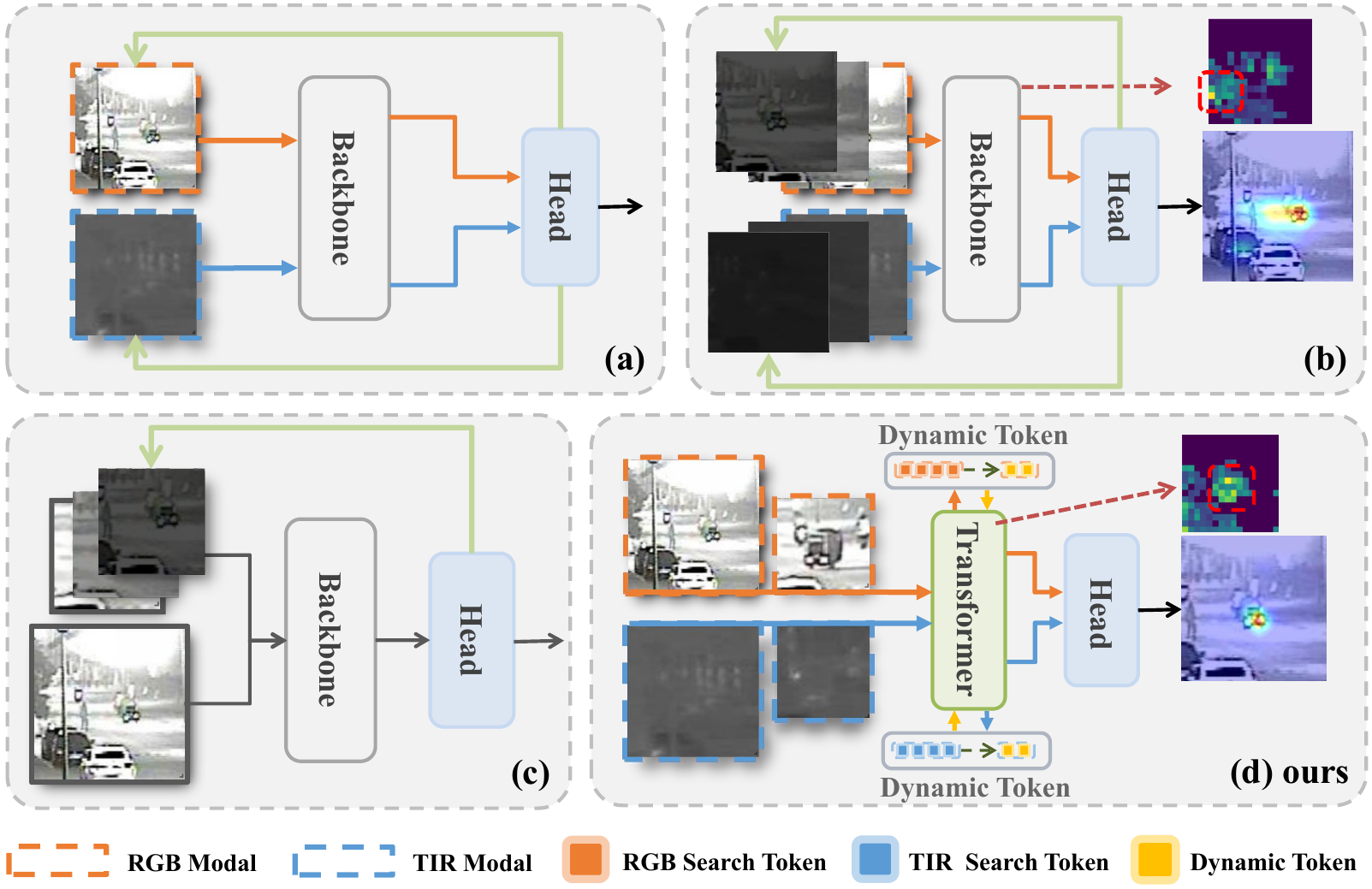}
    \caption{Comparison between our approach and previous ones. (a) Non-temporal method that just focuses on modality fusion. (b) Previous RGBT tracking approaches introduced entire historical frames to incorporate temporal information. (c) The single-modal method introduces temporal information by replacing the initial template. (d) Our approach mixes spatio-temporal multimodal tokens from the static multimodal templates and multimodal search regions in Transformer to handle target appearance changes. }
    \label{fig::motivation}
\end{figure}
Existing RGBT tracking can be divided into two kinds: non-temporal methods and temporal-based methods. Considering that an important point of multi-modal tracking tasks is the fusion and utilization of information among different modalities, traditional non-temporal methods~\cite{li2020challenge,xu2021multimodal,wang2020cross} focus on modal interaction and fusion, and mostly employ MDNet~\cite{nam2016learning} with VGG-M~\cite{simonyan2014very} as their basic tracking network. These CNN-based methods lack global context modeling capability, resulting in limited performance for RGBT tracking. Recently, Transformer has achieved great success in tasks such as natural language modeling~\cite{kenton2019bert,radford2019language} and speech recognition~\cite{luscher2019rwth}, which has led to the emergence of some Transformer-based non-temporal RGBT tracking methods. For example, some works~\cite{feng2022learning,hou2022mirnet} directly fuses the features of two modalities through one or more cross-attention networks. Another work\cite{xiao2022attribute} proposes an attribute-based progressive fusion network for enhancing specific information in modal fusion attribute features. 
However, as shown in Fig.~\ref{fig::motivation}(a), these non-temporal methods fix the initial target appearance as a guide for tracking without making any changes during the target movement and tracking process, so their tracking performance may be seriously degraded over time. 
Although this type of method strengthens the design of modal fusion, they overlook the impact of utilizing temporal information on the robust extraction process of each modality's representation before fusion. 
Target appearance changes can affect the modality feature extraction process, and neglecting temporal information may result in modality features that do not promptly reflect the current target representation.

From the perspective of the tracking task, as the target moves, its appearance in the camera view inevitably deviates from the original appearance information of the target. 
Compared with non-temporal methods, temporal-based methods~\cite{tang2022temporal,zhang2023dual,feng2020learning} show stronger robustness in handling changes in target appearance. 
This is attributed to their ability to take into account historical information, allowing them to better adapt to variations in the target's appearance over time, as shown in Fig.~\ref{fig::motivation}(b).
However, existing temporal methods introduce temporal information by directly fusing features from historical frames with the current frame, and this crude fusion method disrupts the appearance features of the current frame's target. 
These methods inherit the tracking framework of the MDNet~\cite{nam2016learning} network structure, which classifies pixels to distinguish between foreground and background. 
Consequently, when dealing with historical frames where the target is occluded, this fusion method causes the features sent to the localization head to be unable to distinguish between foreground and background. 

Through the above analysis, in multi-modal tracking, the sufficient utilization of temporal information has not received the same attention as modal fusion. 
In contrast, single-modal tracking has also made some progress in this aspect. 
Some works are dedicated to updating the appearance information by replacing or fusing the templates~\cite{borsuk2022fear,zhang2019learning,yang2018learning,zhang2019learning}, as shown in Fig.~\ref{fig::motivation}(c). 
However, the template update strategy of this kind of method is still straightforward and brute-force. In most cases, the initial template is used as a reliable source of target features, so the crude replacement or fusion of the initial template by the dynamic template may destroy the original reliable information, leading to the degradation of the information in the original static template, and continuous accumulation of errors. This seriously affects the tracking performance. Therefore, addressing such issues becomes even more challenging. 

Given the above discussion, in multi-modal tracking, while introducing temporal information to aid in the perception of target variations in the search regions, it is also crucial to retain the original and reliable template information. 
Therefore, we propose a novel scheme to address this issue. We introduce modality enhancement during the joint feature extraction process and incorporate modality-enhanced dynamic tokens to guide the search regions of different modalities to focus on the variations of the target.
The main advantage of this scheme is that it retains the original template as a reliable source of target appearance information, participating in the feature extraction and template matching process for each frame. 
Simultaneously, it embeds dynamic tokens into the process to assist in tracking target changes within the search region.

In order to implement the above scheme, we designed a Spatio-Temporal Multimodal Tokens (STMT) module. 
Concretely, our module first performs modality enhancement of the templates of RGB and TIR modalities by cross-attention, using reliable template information as a medium to convey modal interaction information. 
Similarly, the introduced dynamic tokens also incorporate modality-enhanced operations for processing. 
Cross-attention mechanism~\cite{vaswani2017attention} is a widely adopted and effective practice for context aggregation. 
Its characteristics satisfy the information interaction by receiving information from different sources in an attentional way, so we design cross-attention to handle the current frame search tokens and historical frame search tokens for our specific task. 
We take each modal current search region as a query with more emphasis on current temporal information and take modality-enhanced dynamic tokens for key and value assignment target change information from dynamic tokens to search region. 
Finally, the jointly modality-enhanced templates and temporally fused search regions are passed to the next encoding layer for joint feature extraction and template matching.

Our contributions are summarized as follows:
\begin{itemize}
    \item We propose a novel Spatio-Temporal Multimodal Tokens (STMT) module that incorporates independent dynamic tokens from the past as an additional source of information for target appearance changes which allows the network to focus on target variations without compromising the integrity of the initial target information. It also utilizes modality-enhanced cues from the original template to facilitate hierarchical modality interactions.
    
    \item We extend the ViT architecture with the proposed STMT module to RGBT tracking. To meet the requirements of temporal training, we propose a time-sampling training strategy to achieve a unified RGBT tracking network. 

    \item Extensive experiments demonstrate that our method outperforms other state-of-the-art tracking algorithms on three popular RGBT tracking datasets, and runs at a real-time inference speed of 39.1 FPS.
\end{itemize}

\section{Related Work}
\subsection{RGBT Tracking}
Compared with single-modal tracking tasks, multimodal tracking can use the complementary information of two modalities to target the poor quality of single-modal, and the complementarity of visible and thermal infrared modes is particularly prominent in this regard. 
So, there is an increasing amount of multimodal tracking work focused on the study of RGBT tracking. 
Some of the early work was designed under a network with CNN as the main architecture, mfDiMP~\cite{zhang2019multi} proposes an end-to-end tracking framework for fusing the RGB and TIR modalities in RGBT tracking. 
The relationship between shared information between modalities and modal heterogeneous information is explored by MANet~\cite{manet}. 
They designed a three-way adapter network to extract information from different modalities and share information between modalities. 
\cite{zhang2021jointly} propose a post-fusion method to obtain global and local multi-channel fusion weights taking into account appearance and motion information and dynamically switching appearance and motion cues. 

Multimodal tracking tasks also face a number of unique challenges, and some work has been directed at addressing these challenges. 
CAT~\cite{li2020challenge} addresses some modal-common and modal-specific challenges by designing different branches and aggregating branches together to form more recognizable target representations to address tracking challenges in an adaptive manner. 
Also for the attribute challenge, APFNet~\cite{xiao2022attribute} designed an adaptive fusion model based on attribute aggregation to aggregate all attribute-specific fused features. 
It proposes an attribute-based progressive convergence network that can enhance the modality-specific information in the fused attribute features. 
Transformer-based methods have also emerged in recent years in the field of RGBT tracking and achieved competitive performance. 
Before sending the features to the head network, Feng \emph{et al}.~\cite{feng2022learning} utilizes cross-attention to fuse the features of the two modalities. 
Mei \emph{et al.}~\cite{mei2023differential} introduces self-attention and cross-attention to model different modal information and modal-shared information fusion on the basis of MANet~\cite{manet} networks. 
However, the previous RGBT approach simply introduced an attention mechanism into the CNN architecture. 
The direct fusion of RGB and TIR search regions inevitably introduces background noise and insufficient cross-channel interaction limiting the potential of the transformer module. 

\subsection{Temporal-based Tracking}
In visual object tracking tasks, it is very important to introduce temporal information to focus on target appearance changes over time.
Compared to spatial-only trackers~\cite{bertinetto2016fully,li2018high,liao2020pg,zhang2020ocean}, spatio-temporal trackers make additional use of temporal information to improve the robustness of the tracker. Most existing temporal research is concentrated in the context of single-modal tracking, which can be broadly divided into two categories: gradient-based and gradient-free methods.

Gradient-based update methods~\cite{wang2020tracking,yang2020roam} require gradient computation in the inference process, one of the classical works is MDNet~\cite{mdnet}. However, given that many real devices deploying deep learning in real application scenarios do not support backpropagation, this limits the development of such methods. In contrast, gradient-free methods have greater potential for practical applications. 

In the gradient-free update approach, some work~\cite{yang2018learning,zhang2019learning} is done to adapt to target changes by updating the template with an additional network structure. 
Updatenet~\cite{zhang2019learning} considers earlier historical information, using all the historical templates to fuse into the latest template features. But this approach changes the information and interactions of the original template inevitably accumulating errors with the updates as well. 
Some Transformer-based work has also emerged in recent years. Stark~\cite{yan2021learning} used the structure of the transformer to process the features of the search region and templates to get the tracking results, while using the historical templates to join them. 
MixFormer~\cite{cui2022mixformer} scores historical frames using the training quality branch which has the disadvantage of secondary training and needs to restart the training quality branch after the main body of the network has been trained. 
These methods all have relatively obvious limitations. 
In most cases, the initial template is used as a reliable source of target features. 
Replacing or fusing dynamic templates in a crude manner will disrupt the integrity of the original reliable information and introduce accumulating errors over time. 
In addition, many template-based temporal methods require the incorporation of additional networks to achieve their functionality. 

In multimodal tasks, the introduction of temporal information and the fusion of modal information is equally important. By combining these two aspects, we can achieve improved performance and robustness in multimodal tasks. 
DMSTM~\cite{zhang2023dual} through the space-time memory reader with bimodal fusion to aggregate features of historical and current frames to share information in the time domain. 
TAAT~\cite{tang2022temporal} introduces an extra search sample adjacent to the original one selected to predict the temporal transformation during the process of picking up template-search image pairs. 
However, the current exploration of temporal aspects in the field of multimodal tracking is very limited. To alleviate these limitations, we propose a Spatio-Temporal Multimodal Tokens module and abandon the various operations of previous temporal methods for updating the original template. We introduce dynamic tokens as supplementary information to assist the search region in focusing on target changes, without compromising the robustness of the interaction between the original template and the search region.

\begin{figure*}[t]
    \centering
    \includegraphics[width=1.0\textwidth]{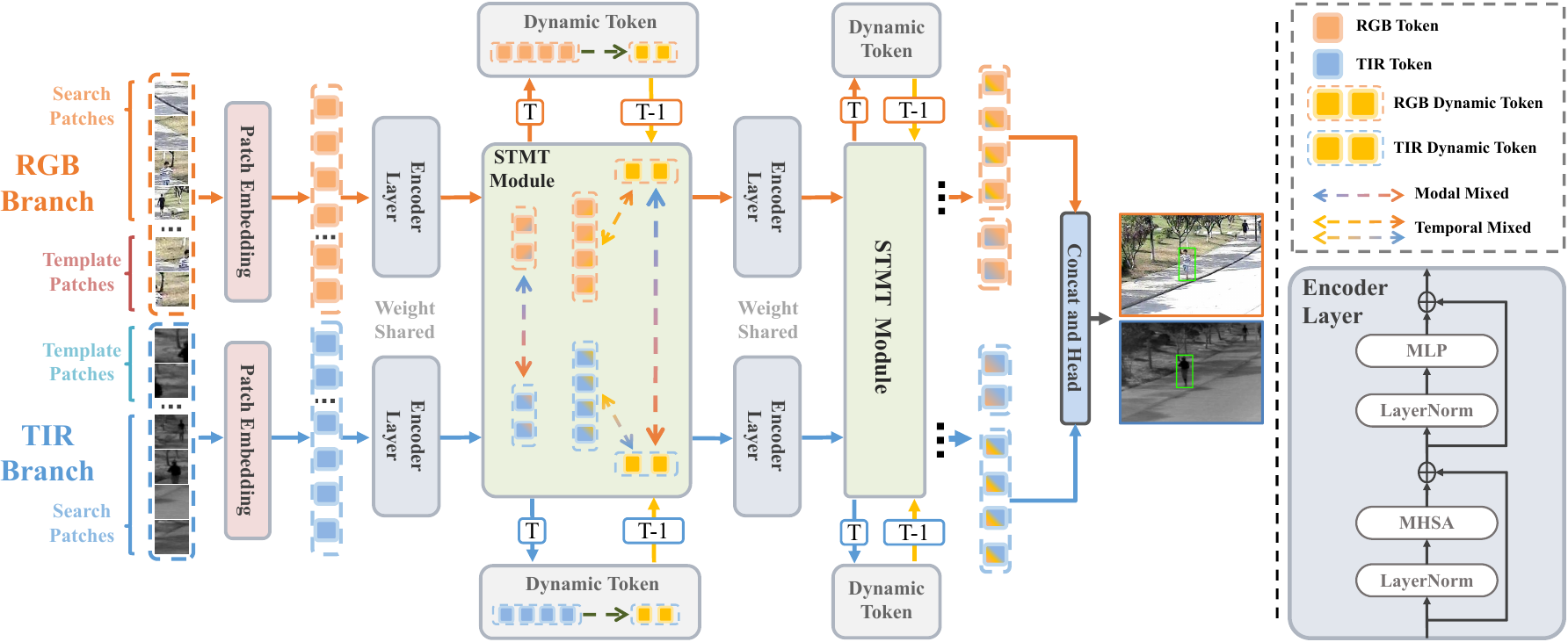} 
    \caption{The overall framework of our proposed Spatio-Temporal Multimodal Tokens Transformer framework for RGBT Tracking. RGB and TIR image patches are embedded as tokens and fed into Transformer blocks for joint feature extraction and intra-modal search-template matching. In the proposed module, T represents the current frame time, and T-1 represents the time of the previous frame. We first extract the search regions of both modalities and form them into dynamic tokens for the next time step. Then, we perform modality enhancement on the static reliable templates to provide modality interaction cues in the subsequent encoding layers for joint feature extraction. Simultaneously, we integrate the dynamic tokens that from the previous time step into the current search region to provide information about target variations.}
    \label{fig::overall}
\end{figure*}

\section{Methodology}
\label{sec:method}
In this section, we propose the Spatio-Temporal Multimodal Tokens Transformer Network. For the sake of clarity, we first introduce the input data format of the multimodal ViT Tracker for RGBT Tracking. Then, the design of the Spatio-Temporal Multimodal Tokens module is explained in detail, including the  extraction of dynamic tokens. In addition, we use a temporal sampling approach to satisfy the training with a uniform network structure. Finally, the overall framework of our method is shown in Fig.~\ref{fig::overall}. 

\subsection{Input of Multimodal ViT Tracker}
\label{sec:mmvt}
In this work, we follow a mainstream tracking framework, that is the siamese framework for multimodal-based tracking. Considering that the structure of ViT \cite{dosovitskiy2020image} meets our multi-tiered requirements in terms of network, we chose OSTrack~\cite{ye2022joint}, which based on ViT as the baseline architecture, The self-attention~\cite{vaswani2017attention} operation helps us integrate information from different sources. In addition, using the publicly available pre-training model provided by OSTrack can significantly reduce our training costs.

Given a ``template-search'' image pair $\{z_v, x_v\}$  in the visible modal and a ``template-search'' image pair $\{z_t, x_t\}$ in the thermal infrared modal, we first crop and resize the template image and search image into $H_{z} \times W_{z} = 128\times 128$ and $H_{x} \times W_{x} = 
256\times 256$, respectively. Here we do not make a modal distinction in the subscripts because we have the same pre-processing operations for the images. Take the visible modal as an example, we split and flatten the template image $z_v$ and the search image $x_v$ into $z_{p_v} \in \mathbb{R}^{N_{z} \times (3 \cdot P^2)}$ and $x_{p_v} \in \mathbb{R}^{N_{x} \times (3 \cdot P^2)}$, where $P^2$ is the resolution of each patch, which we set by default to 16, $N_{z} = H_{z}\times W_{z}/P^2$ and $N_{x} = H_{x}\times W_{x}/P^2$ are the numbers of patches of template and search region respectively.

After that, same as the ViT, a trainable linear projection layer with parameter $E\in\mathbb{R}^{(3 \cdot P^2) \times D}$ is used to project $z_{p_v}$ and $x_{p_v}$ into $D$ dimension latent space. This projection is commonly called Patch Embeddings~\cite{dosovitskiy2020image}. 
Then, two learnable position embeddings ${P_z}\in\mathbb{R}^{N_{z} \times D}$ and $P_x\in\mathbb{R}^{N_{x} \times D}$ are added to the patch embeddings of the template and search region separately to produce the initial template token embeddings $H^0_{z_v}$ and search region token embeddings $H^0_{x_v}$ in visible modal as in Eq.~\ref{eq:init-v}. With the same operation, we obtain the $H^0_{z_t}$ and $H^0_{x_t}$ in the thermal infrared modal as in Eq.~\ref{eq:init-t}. 
\begin{align}
    \label{eq:init-v}
    H^0_{z_v} &= z_{p_v}E + {P_z}, & H^0_{x_v} &= x_{p_v}E + {P_x}\\
    \label{eq:init-t}
    H^0_{z_t} &= z_{p_t}E + {P_z}, & H^0_{x_t} &= x_{p_t}E + {P_x}
\end{align}

Notably, the learnable position embeddings are shared in multimodal. Based on it, we inherit joint feature extraction and search template matching, so finally concatenate them together to get the token sequence inputs $H^0_v$ and $H^0_t$ for visible and thermal infrared modalities, as in Eq.~\ref{eq:seq-vt}. 
Our framework retains the structure of the original 12 Encoder Layers, our module has been inserted between some of the coding layers, the details are described in detail in the following text. 
\begin{align}
    \label{eq:seq-vt}
    H^0_v &= Concat(H^0_{z_v},H^0_{x_v}), & H^0_t &= Concat(H^0_{z_t},H^0_{x_t})
\end{align}

\subsection{Spatio-Temporal Multimodal Tokens}
\label{sec:met}
In the previous temporal-based approaches, the crude replacement or fusion of the initial template by the dynamic template may damage the robustness of the search template match seriously, and cause continuous accumulation of errors. To address these limitations, we design a novel module that introduces dynamic tokens to model temporal information fusion capabilities. It's worth noting that, unlike traditional single-modal tasks, in our task, we must also consider modality interactions and utilize the complementary information between modalities to enhance each other. The implemented details are as follows. \\

\noindent\textbf{Modality Enhancement Interaction}

\begin{figure}[t]
    \centering
    \includegraphics[width=1.0\linewidth]{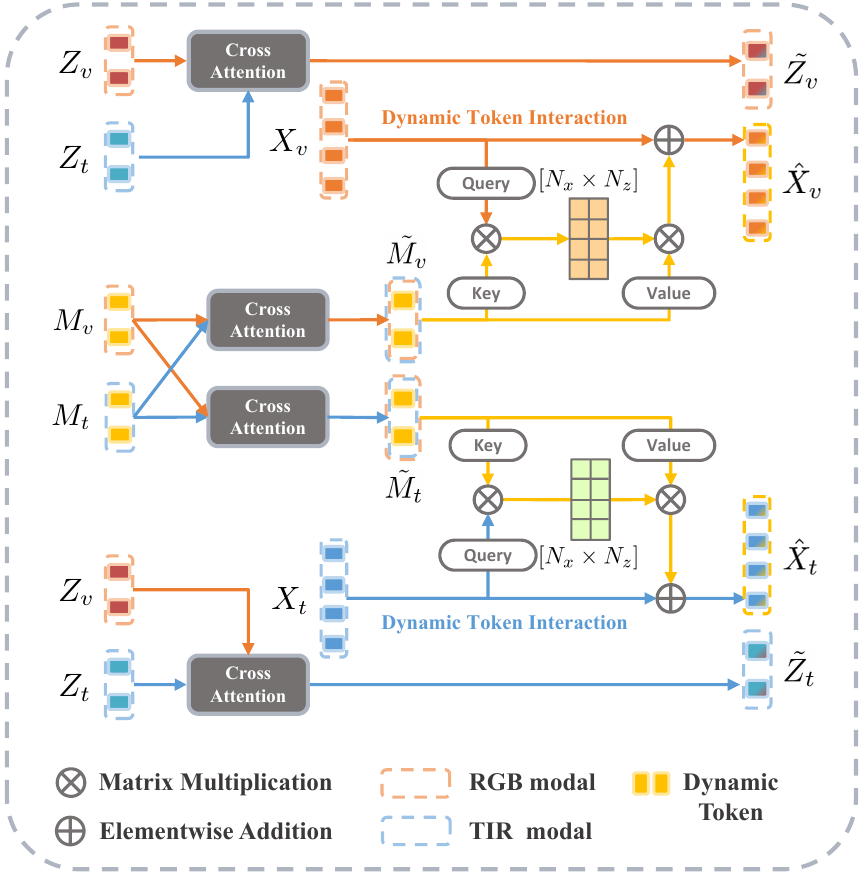} 
    \caption{Conceptual illustration of Spatio-Temporal Multimodal Tokens (STMT) module. For clarity, we only present the core design aspects and omit details such as template updates and operations like LN and MLP.}
    \label{fig::met}
\end{figure}

Given the output of an encoder layer immediately preceding the layer in which we want to embed the module, that is a pair of token sequences $H^{i}_v, H^{i}_t$ from two modalities. The core design of the module is illustrated in Fig.~\ref{fig::met}. First, we split the two token sequences $H^{i}_{v}$ and $H^{i}_{t}$ respectively to get the corresponding template and search parts as in Eq.~\ref{eq::token},
\begin{align}
    \label{eq::token}
    H^{i}_v=\{Z_v,X_v\} & , & H^{i}_t=\{Z_t,X_t\},
\end{align}
where $i$ is the encoder layer sequence number. The $Z$ and $X$ denote tokens belonging to the template and search regions, and the subscripts $v$ and $t$ indicate the visible modal and thermal infrared modal respectively. 

Then we perform a modality enhancement interaction for each modality. Take the visible modal as an example, the attention matrix transforms the visible modal template tokens to obtain the Query, and the thermal infrared modal template tokens are generated and provided with the Key and Value in attention operation. We send them to the cross-attention module to obtain modal-mixed template tokens from TIR to RGB modalities as in Eq.~\ref{eq:stv-cross}. It is worth noting that the TIR and RGB share the same modality enhancement parameters. Through experimental observations, we have found that sharing parameters in modality fusion attention leads to improved learning of the fusion patterns between the two modalities. 
\begin{align}
    \label{eq:stv-cross} G_{Z_{vt}}=Softmax\left(\frac{Z_{vQ} Z_{tK}^{\top}}{\sqrt{C}}\right) \cdot Z_{tV},
\end{align}
where the $Z_{vQ},Z_{tK},Z_{tV}$ denote the query $Q$, key $K$, and value $V$. $C$ obtained by performing projections on $Z_v$ and $Z_t$. $C$ is the number of feature channels. 
Both modalities have the same feature dimension, which is achieved through data processing techniques. Then, the TIR-relevant context $G_{Z_{vt}}$ is refined and integrated with $Z_{v}$ to enrich the RGB template tokens with the required TIR information:
\begin{align}
    \label{eq:stv-cross-more}
    {Z^{'}}_v &= Z_v + G_{Z_{vt}}, &
    \tilde{Z_v} &= {Z^{'}}_v + MLP(LN({Z^{'}}_v)),
\end{align}
where $LN$ and $MLP$ represent LayerNorm and Multilayer Perceptron. 

Similarly, for the thermal infrared modal, the operation is symmetrical, with the thermal infrared modal providing the Query and the visible modal providing the Key and Value. By integrating the RGB-relevant context $G_{Z_{tv}}$, we can get the modal-mixed template tokens for the thermal modal:
\begin{align}
    \label{eq:svt-cross-more}
    {Z^{'}}_t &= Z_t + G_{Z_{tv}}, &
    \tilde{Z_t} &= {Z^{'}}_t + MLP(LN({Z^{'}}_t)),
\end{align}
to simplify, we denote the cross-modal enhancement attention above as $CA(I_q, I_{kv})$, so the aforementioned process can be rewritten as:
\begin{align}
   \label{eq:stt-cross}
    \tilde{Z_v}=CA(Z_v,Z_t), &&
    \tilde{Z_t}=CA(Z_t,Z_v) 
\end{align}

Considering the appearance changes of moving targets, we further propose dynamic tokens in addition to templates. 
The dynamic tokens are also a token sequence and are the same length as the template tokens. Further details about extracting dynamic tokens can be found in Section.~\ref{sec:thfm}. Here, similar to the operation on templates, We also perform modality enhancement to obtain the modal-mixed dynamic tokens:
\begin{align}
    \label{eq:dt-cross}
    \tilde{M_v}=CA(M_v,M_t), &&
    \tilde{M_t}=CA(M_t,M_v)
\end{align}
where $M$ denotes dynamic tokens, and the subscript indicates which modal it comes from. \\

\noindent\textbf{Dynamic Token Interaction}

After that, we make the modal-mixed dynamic tokens work separately for each modal search region. 
Taking visible modal as an example again, visible search region tokens serve as query and the dynamic tokens serve as key and value to distribute information via similar cross-modal attention:
\begin{equation}
    \label{eq:temporal-attention}
    G_{Xvm}=\operatorname{Softmax}\left(\frac{(X_v W_Q) (\tilde{M_v} W_K)^{\top}}{\sqrt{C}}\right) \cdot (\tilde{M_v} W_V),
\end{equation}
\begin{equation}
    \label{eq:temporal-attention-more}  
    {X^{'}}_v = X_v + G_{Xvm},
\end{equation}
\begin{equation}
    \hat{X}_v = {X^{'}}_v + MLP(LN({X^{'}}_v)),
\end{equation}
where $W_Q, W_K, W_V$ denote parameters of the query, key, and value projection layers. 
Similarly, TIR search region tokens interact with their modal-mixed dynamic tokens to obtain $\hat{X}_{t}$. 
To simplify, we denote the interaction between the search region and the modal-mixed dynamic tokens above as $TF(X,\tilde{M})$, so we can rewrite the aforementioned process as:
\begin{equation}
    \label{eq::temporal-fution} 
    \hat{X}_v=TF(X_v,\tilde{M_v}),\quad \hat{X}_t=TF(X_t,\tilde{M_t}),
\end{equation}

Finally, we concatenate each modal-mixed template tokens with the spatio-temporal multimodal search tokens to obtain the input for the next Transformer Block $\hat{H}^{i}_v$ and $\hat{H}^{i}_t$, and send them to the next Encoder Layer:
\begin{align}
    \label{eq::concat}
    \hat{H}^{i}_v &= Cat(\tilde{Z_v},\hat{X}_v), &
    \hat{H}^{i}_t &= Cat(\tilde{Z_t},\hat{X}_t),\\
    \label{eq::encoder}
    H^{i+1}_v &= EL^{i+1}(\hat{H}^{i}_v), &
    H^{i+1}_t &= EL^{i+1}(\hat{H}^{i}_t),
\end{align}
where $EL(\cdot)$ denotes the input passing through an encoder layer and undergoing a series of operations to yield the subsequent layer's output. 
Thus, the search region jointly completes the aggregation of modal-mixed dynamic tokens, which capture the latest target feature information. Then, it is combined with the modal-mixed template tokens. Through consecutive Transformer Blocks, features from the search and template region tokens are extracted step by step, capturing their matching relationships and enhancing the target localization capabilities in each modality separately. In addition, by sharing the parameters of the two modal Encoder Layers, we can not only facilitate the learning of shared feature information between the two modalities but also avoid redundancy.

\subsection{Multimodal Dynamic Token}
\label{sec:thfm}
In the previous subsection, we introduced the Spatio-Temporal Multimodal Tokens module in detail, which includes modality enhancement interaction and dynamic token interaction. Among them, a key point is how to obtain the Multimodal Dynamic Tokens. Here, we elaborate on this procedure, which is also one of the main contributions of this paper. Unlike the temporal-based methods with template updating, our objective is not only to utilize temporal information to focus on target changes but also to preserve the original reliable information of the initial template and ensure robust interaction with the search region is not compromised. 
Therefore, we propose the Multimodal Dynamic Tokens to introduce independent temporal information for enhancing the focus of the search region on target variations. 

Specifically, we start by splitting the joint tokens into template tokens and search tokens. 
Due to the elimination mechanism of OStrack\cite{ye2022joint}, the search tokens at this point may not have the original length. 
Therefore, we restore them to their original length based on the saved eliminated token indices. 
The content of the eliminated tokens is filled with zeros. 
Finally, we save the processed search tokens into the multimodal search token collection $M_{thf}$:
\begin{align}
    \label{eq:Mthf}
   M_{thf}=\left\{(H^i_{X_v},H^i_{X_t}),(H^j_{X_v},H^j_{X_t}),\cdots\right\}
\end{align}
where $i, j$ denote the sequential indices of the token hierarchies that we want to preserve, and the subscripts $X_v, X_t$ represent the search region from two modalities. 

\begin{figure}[t]
    \centering
    \includegraphics[width=1\linewidth]{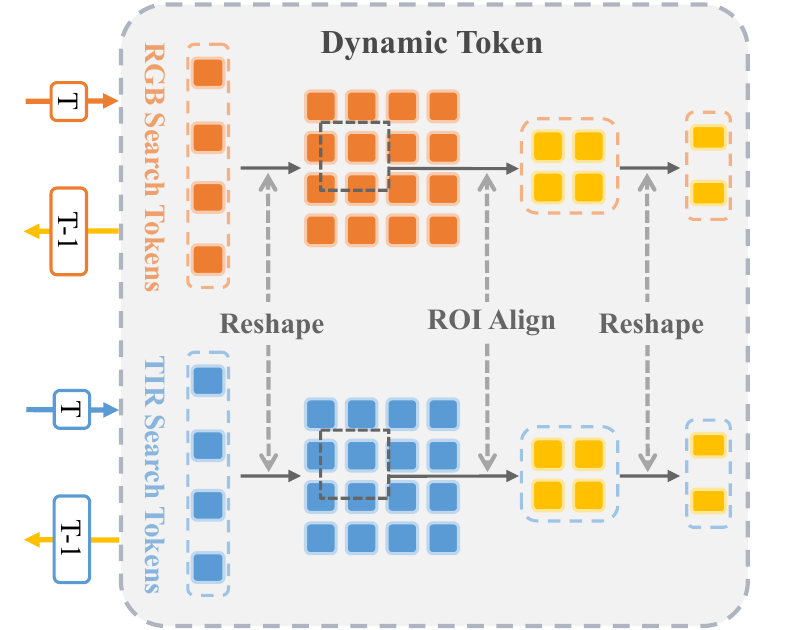} 
    \caption{The extraction process of multimodal dynamic tokens is demonstrated. The current frame at the current time (T) step is passed through a process of reshaping, ROI cropping, and another reshaping to obtain dynamic tokens. These dynamic tokens are preserved for the next time step, and the dynamic tokens from the previous time step (T-1) are also input to the network for the current time step.}
    \label{fig::dynamictoken}
\end{figure}

During the inference process, after completing the tracking procedure for the current frame, we obtain the confidence score and bounding box of the target localization from the output of the $Head$. 
The confidence score serves as the basis for determining whether to update. 
When the specified update interval is reached, and the confidence score exceeds the threshold we set, we process the saved hierarchical multimodal search tokens. 
As shown in Fig.~\ref{fig::dynamictoken}, we reverse the process of patch embedding to reshape the hierarchical search tokens into 2D feature maps. 
Based on the bounding box results, we apply the ROI Align technique~\cite{he2017mask} to crop the feature maps in the feature dimension, obtaining dynamic features with the same dimensions and size as the template features. Next, we reshape them into token form and replace the original cache, saving it to the multimodal dynamic token collection: 
\begin{equation}
    M_v =R\left(ROI\left(R(H_{X_v})\right)\right),
\end{equation}
\begin{equation}
    M_t =R\left(ROI\left(R(H_{X_t})\right)\right),
\end{equation}
\begin{equation}
    M_{thf}=\left\{(M^i_v,M^i_t),(M^j_v,M^j_t),\cdots\right\},
\end{equation}
where $R(\cdot)$ denotes reshape operation, $ROI(\cdot)$ denotes the ROI Align~\cite{he2017mask} operation. 
When the update condition is met, the multimodal dynamic tokens are passed into the tracking process of the next frame. 
Through the method mentioned earlier, the temporal information assists in obtaining better target variation information for subsequent tracking.

\subsection{Temporal Training}
Since our module focuses on target changes by introducing the temporal information independent of the initial template, the training of the proposed module requires additional time series information. However, traditional training methods alone cannot provide this information. Some temporal-based approaches first output a set of results from the tracker and then use these results and additional networks to train their temporal modules.
However, this inevitably introduces additional cumbersome steps into the process. 

In contrast, our temporal module is embedded in the feature extraction, and we aim to achieve a complete network that can be directly used in the inference stage through a single training process without additional network architectures or designs. Therefore, we employ a temporal sampling strategy, which allows the training stage to focus on tracking, localization, and classification training and also enables training of our temporal module. 
Specifically, during the sampling process, we typically randomly select a dataset and randomly choose a sequence from it. 
Then, within the sequence, we randomly sample a frame as the template region and select a frame different from the current frame as the search region. 
These paired sampled data $S=\{S^x_v,S^z_v,S^x_t,S^z_t\}$ are then used for training. 
At the same time, we also sample two frames from the current sequence that are different from the previous pair of data. 
These frames are used to create template and search regions, forming another pair of temporal samples $T=\{T^x_v, T^z_v, T^x_t, T^z_t\}$ that are included in the training process. 

Next, the temporal samples are fed into the network backbone for feature extraction. 
At the hierarchical level where we need to embed the module, we extract the temporally sampled template tokens to simulate the dynamic tokens used during the actual inference process.
Finally, these dynamic tokens are combined with the first set of samples $S$ and fed back into the network for temporal training. 

\section{Experiments}
In this section, we evaluate our algorithm by comparing the tracking performance with some state-of-the-art trackers on three RGBT tracking benchmarks, including RGBT210, RGBT234, and LasHeR. Based on the experiment results, we validate the effectiveness of the proposed method and analyze the major components of the algorithm. Our tracker is implemented in PyTorch 1.7.1, python 3.8, CUDA 12.1 and runs on a computer with 4 GeForce RTX 3090 GPU cards.

\subsection{Datasets and Evaluation Metrics}
We first introduce the details of the datasets and the evaluation metrics.
We use three large RGBT tracking datasets. 
\textbf{RGBT210}~\cite{li2017weighted} is the first large-scale RGBT dataset, which contains 210 video sequence pairs, 210K frames, and 12 tracking challenge attributes. \textbf{RGBT234}~\cite{li2019rgb} is a larger RGBT tracking dataset than GTOT~\cite{li2016learning}, which is extended from the RGBT210~\cite{} dataset and provides more accurate annotations that take into full consideration of various environmental challenges. 
Contains 234 RGBT highly aligned video pairs with about 234K frames in total, and 12 attributes are annotated to facilitate analyzing the effectiveness of different tracking algorithms for different challenges. 
\textbf{LasHeR}~\cite{li2021lasher} is the largest RGBT tracking dataset at present, which contains 1224 aligned video sequences including more diverse attribute annotations, in which 245 sequences are divided separately as testing datasets, and the remaining are designed for training datasets.

We adopt the precision rate (PR) and success rate (SR) in the one-pass evaluation (OPE) as evaluation metrics for quantitative performance evaluation. 
Herein, PR measures the percentage of all frames whose distance between the center point of the tracking result and ground truth is less than the threshold, and we compute the representative PR score by setting the threshold to 20 pixels in three datasets. 
SR measures the percentage of successfully tracked frames whose overlaps are larger than thresholds, and we calculate the representative SR score by the area under the curve.

\begin{table*}[htb]
	\centering
    \setlength{\tabcolsep}{0.22cm}
    \renewcommand\arraystretch{1.3}	
	\caption{PR, NPR, and SR scores (\%) of our traker on RGBT210, RGBT234 and the testing set of LasHeR against other trackers. The best and second results are in $\color{red} red$ and $\color{blue} blue$ colors, respectively. * indicates the tracker is re-trained.}
	\begin{tabular}{cccccccccccc}
		\hline
        \hline
		\multirow{2}{*}{Methods} &\multirow{2}{*}{Pub. Info.} &\multirow{2}{*}{Framework}&\multirow{2}{*}{is-Temporal}& \multicolumn{2}{c}{RGBT210} & \multicolumn{2}{c}{RGBT234} & \multicolumn{3}{c}{LasHeR} & \multirow{2}{*}{FPS$\uparrow$} \\
		&&& & MPR$\uparrow$ & MSR$\uparrow$ & MPR$\uparrow$ & MSR$\uparrow$ & PR$\uparrow$ & NPR$\uparrow$ & SR$\uparrow$ & \\
		\hline
		DAPNet~\cite{zhu2019dense} & ACM MM 2019 &CNN &\XSolidBrush & - & - & 76.6 & 53.7 & 43.1 & 38.3 & 31.4 & 2\\
        MANet~\cite{manet} & ICCVW 2019 &CNN &\XSolidBrush & - & - & 77.7 & 53.9 & - & - & - & 1.1\\
		DAFNet~\cite{gao2019deep} & ICCVW 2019 &CNN &\XSolidBrush & - & - & 79.6 & 54.4 & 44.8 & 39.0 & 31.1 & 20\\
		mfDiMP~\cite{zhang2019multi} & ICCVW 2019 &CNN &\XSolidBrush & 78.6 & 55.5 & - & - & 44.7 & 39.5 & 34.3 & 10.3 \\
  	FSRPN~\cite{kristan2019seventh} & ICCVW 2019 &CNN &\XSolidBrush & 68.9 & 49.6 & 71.9 & 52.5 &- &- &-& 29 \\
		CMPP~\cite{wang2020cross} & CVPR 2020  &CNN &\XSolidBrush & - & - & 82.3 & 57.5 & - & - & - &1.3 \\
		MaCNet~\cite{zhang2020object} & Sensors 2020  &CNN &\XSolidBrush & - & - & 79.0 & 55.4 & 48.2 & 42.0 & 35.0 & 0.8\\
		CAT~\cite{li2020challenge} & ECCV 2020  &CNN &\XSolidBrush & 79.2 & 53.3 & 80.4 & 56.1 & 45.0 & 39.5 & 31.4 & 20\\
		FANet~\cite{zhu2020quality} & TIV 2021  &CNN &\XSolidBrush & - & - & 78.7 & 55.3 & 44.1 & 38.4 & 30.9 & 19\\
		M5L~\cite{tu2021m} & TIP 2021  &CNN &\XSolidBrush & - & - & 79.5 & 54.2 & - & - & - & 9.7\\
		ADRNet~\cite{zhang2021learning} & IJCV 2021 &CNN &\XSolidBrush & - & - & 80.7 & 57.0 & - & - & - & 25\\
		JMMAC~\cite{zhang2021jointly} & TIP 2021 &CNN &\XSolidBrush & - & - & 79.0 & 57.3 & - & - & - & 4\\
		MANet++~\cite{lu2021rgbt} & TIP 2021  &CNN &\XSolidBrush & - & - & 80.0 & 55.4 & 46.7 & 40.4 & 31.4 & 25.4 \\
		DMCNet~\cite{lu2022duality} & TNNLS 2022 &CNN &\XSolidBrush & 79.7 & 55.5 & \color{blue}83.9 & 59.3 & 49.0 & 43.1 & 35.5 &2.3 \\
		TFNet~\cite{zhu2021rgbt} &TCSVT 2022 &CNN &\XSolidBrush & 77.7 & 52.9 &80.6 & 56.0 &- &- &-& 17 \\		
		HMFT~\cite{zhang2022visible}& CVPR 2022 &CNN &\XSolidBrush & 78.6 & 53.5 & 78.8 & 56.8 & - & - & - & 30.2 \\
  	APFNet~\cite{xiao2022attribute} & AAAI 2022 &CNN+Trans &\XSolidBrush & - & - & 82.7 & 57.9 & 50.0 & 43.9 & 36.2 & 1.3\\
		MIRNET~\cite{hou2022mirnet} & ICME 2023 &CNN+Trans &\XSolidBrush & - & - & 81.6 & 58.9 & -& - & - &30 \\    
        ViPT~\cite{zhu2023visual} & CVPR 2023        &Trans &\XSolidBrush &- & - & 83.5 & \color{blue}61.7 & \color{blue}65.1 & - & \color{blue}52.5 & -\\
        DRGCNet~\cite{mei2023differential} & IEEE SENS J 2023 &CNN+Trans &\XSolidBrush & - & - & 82.7 & 58.1 & 48.3 & 42.3 & 33.8 &4.9 \\
		\hline	
        CCFT~\cite{feng2020learning} & VCIR 2020 &CNN &\Checkmark &76.0 & 54.6 & 78.3 & 58.1 & - & - & - & 12.6\\
        TAAT~\cite{tang2022temporal} & - &CNN &\Checkmark &71.0 & 48.6 & - & - & - & - & - & -\\
        DMSTM~\cite{zhang2023dual} & TIM 2023 &CNN &\Checkmark &- & - & 78.6 & 56.2 & 55.7 & 50.3 & 40.0 & 27.6\\
        \hline	
        OSTrack~\cite{ye2022joint}-RGBT & - &Trans &\XSolidBrush &75.8 & 55.3 & 78.6 & 59.1 & 53.0 & 50.1 & 43.0 & \color{red}45.5\\
        OSTrack~\cite{ye2022joint}-RGBT* & - &Trans &\XSolidBrush &\color{blue}81.3 & \color{blue}57.7 & 83.0 & 61.5 & 64.1 & \color{blue}60.2 & 51.1 & \color{red}45.5\\
        \hline
		\bf STMT & -&Trans &\Checkmark & \color{red}83.0 & \color{red}59.5 & \color{red}86.5 & \color{red}63.8 & \color{red}67.4 & \color{red}63.4 &\color{red}53.7 & \color{blue}39.1\\
		\hline
	\end{tabular}
	\label{tb::result}
\end{table*}

\subsection{Implementation Details}
We set the training batch size to 12, We train our model for 60 epochs on the LasHeR dataset with 60k image pairs per epoch and directly evaluate our model on RGBT234, RGBT210 and LasHeR datasets without further finetuning. 
The learning rate is set as 1e-6 for the backbone, 1e-4 for our module parameters, and 1e-5 for the head parameters, which is decayed by 10× after 20 epochs. 
We adopt AdamW~\cite{loshchilov2017decoupled} as the optimizer with 1e-4 weight decay. 
The search regions are resized to 256 × 256 and templates are resized to 128 × 128. Our module is inserted in the 4-th, 7-th, and 10-th layers of the baseline. 

\subsection{Quantitative Comparison}
We test our method on three popular RGBT tracking benchmarks, including RGBT210~\cite{li2017weighted}, RGBT234~\cite{li2019rgb}, and LasHeR~\cite{li2021lasher}, and we compare performance with some state-of-the-art trackers which could be divided into two categories. 
First, some methods that emphasize modality interaction design include DAPNet~\cite{zhu2019dense}, MANet~\cite{manet}, DAFNet~\cite{gao2019deep}, mfDiMP~\cite{zhang2019multi}, FSRPN~\cite{kristan2019seventh}, CMPP~\cite{wang2020cross}, MaCNet~\cite{zhang2020object}, CAT~\cite{li2020challenge}, FANet~\cite{zhu2020quality}, M5L~\cite{tu2021m}, ADRNet~\cite{zhang2021learning}, JMMAC~\cite{zhang2021jointly}, MANet++~\cite{lu2021rgbt}, DMCNet~\cite{lu2022duality}, TFNet~\cite{zhu2021rgbt}, HMFT~\cite{zhang2022visible}, APFNet~\cite{xiao2022attribute}, MIRNET~\cite{hou2022mirnet}, ViPT~\cite{zhu2023visual} and DRGCNet~\cite{mei2023differential}.

In addition, we also compare with some temporal-based methods such as CCFT~\cite{feng2020learning}, TAAT~\cite{tang2022temporal} and DMSTM~\cite{zhang2023dual}.
\begin{figure}[t]
\centering
\includegraphics[width=0.5\textwidth]{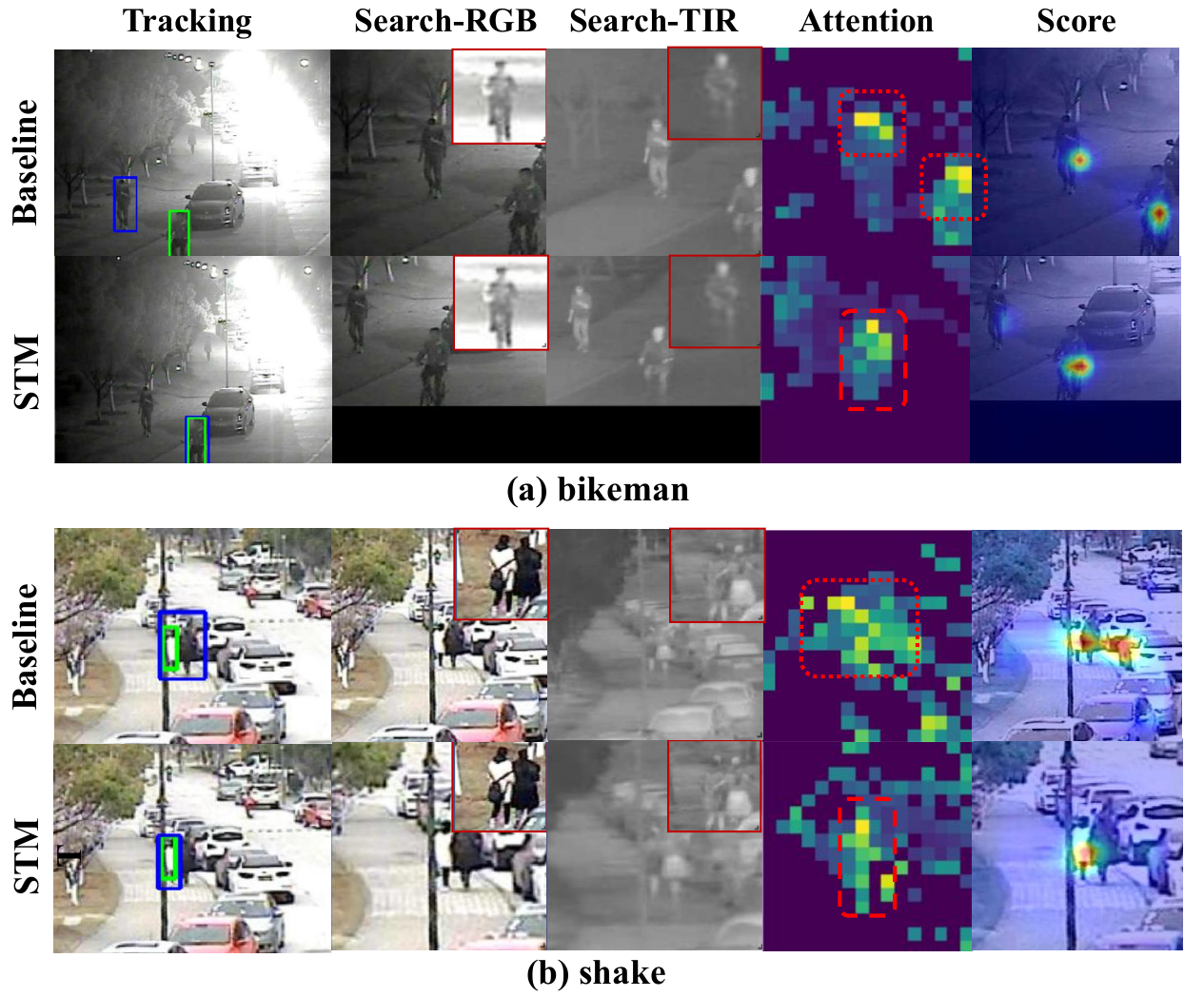} \\
\caption{Some visual cases of tracking result on RGBT234. It shows the comparison between STMT and the baseline is demonstrated on two sequences, where the blue tracking boxes represent the results of the tracker, while the green boxes indicate the ground truth.}
\label{fig::metVSbase}
\end{figure}

\subsubsection{Evaluation on RGBT210 dataset}
As shown in Table~\ref{tb::result}, we can see that the performance of our tracker is clearly superior to the state-of-the-art RGBT methods in all metrics. 
In particular, the PR/SR score of our method is 3.3\%/4.0\% higher than that of the best non-temporal method DMCNet~\cite{lu2022duality}. Compared with our baseline method, the MPR (Mean Precision Rate)/MSR (Mean Success Rate) scores of our tracker are improved by 1.7\%/1.8\%, which is sufficient to prove the effectiveness and superiority of our method on RGBT210.  

\begin{table*}[htp]	
\setlength{\tabcolsep}{0.5cm}
\centering
\renewcommand\arraystretch{1.5}	
\caption{The tracking results (PR/SR) under each attribute on RGBT234 dataset. (The top two results are highlighted in \textcolor{red}{red} and \textcolor{blue}{blue}, respectively). } 
\begin{tabular}{c|cccccc|c}
\hline 
\hline 
\textbf{Attribute}   &\textbf{MANet++}   &\textbf{TFNet} &\textbf{APFNet}  &\textbf{DMCNet}                                  &\textbf{DMSTM}                   &\textbf{OSTrack-RGBT*}                                          &\textbf{\bf STMT} \\
\hline 
\textbf{PO}      &81.4/56.8          &83.6/57.8      &86.3/60.6        &\textcolor{red}{89.5}/\textcolor{blue}{63.1}          &{79.0}/{57.2}                               &{80.2}/{58.5}                      &\textcolor{blue}{88.9}/\textcolor{red}{66.2} \\
\textbf{TC}      &76.1/54.8            &80.9/57.7      &82.2/58.1        &\textcolor{blue}{87.2}/\textcolor{blue}{61.2}       &{72.2}/{50.7}                             &{80.4}/{58.2}                   &\textcolor{red}{87.6}/\textcolor{red}{65.0} \\
\textbf{BC}        &75.2/47.2        &81.3/52.5      &81.3/54.5        &{83.8}/{55.9}                                     &{72.4}/{47.9}                  &\textcolor{blue}{84.6}/\textcolor{blue}{62.0}                    &\textcolor{red}{85.4}/\textcolor{red}{64.3} \\
\textbf{MB}      &72.0/50.1           &70.2/50.6      &74.5/54.5        &{77.3}/{55.9}                                 &{72.9}/{52.7}                            &\textcolor{red}{86.5}/\textcolor{blue}{63.4}           &\textcolor{blue}{85.4}/\textcolor{red}{64.0} \\
\textbf{DEF}     &77.7/54.2          &76.5/54.3      &78.5/56.4        &77.9/56.5                                        &{80.0}/{58.9}                   &\textcolor{blue}{84.3}/\textcolor{blue}{62.0}                             &\textcolor{red}{89.0}/\textcolor{red}{64.7} \\
\textbf{LI}     &77.2/51.1           &80.5/54.1      &84.3/56.9        &\textcolor{blue}{85.3}/{58.7}            &{78.8}/{55.7}                              &84.4/\textcolor{blue}{61.0}                             &\textcolor{red}{90.6}/\textcolor{red}{68.9}\\
\textbf{CM}       &72.9/50.3         &75.0/53.4      &77.9/56.3        &{80.1}/{57.6}                            &{76.5}/{54.8}                              &\textcolor{blue}{84.8}/\textcolor{blue}{63.2}                   &\textcolor{red}{88.3}/\textcolor{red}{65.5} \\
\textbf{SV}      &78.1/55.5            &80.3/56.8      &83.1/57.9        &{84.6}/59.8                               &{84.0}/{61.1}           &\textcolor{blue}{84.9}/\textcolor{red}{63.4}                   &\textcolor{red}{86.6}/\textcolor{blue}{62.9} \\
\textbf{FM}       &67.8/43.8            &78.2/49.0      &79.1/51.1        &{80.0}/{52.4}     &{76.0}/{51.6}                          &\textcolor{red}{85.2}/\textcolor{red}{62.8}                                       &\textcolor{blue}{80.6}/\textcolor{blue}{57.1} \\
\textbf{LR}       &77.8/50.5          &83.7/54.4      &{84.4}/56.5        &85.4/{57.9}           &{75.2}/{51.0}                       &\textcolor{red}{88.1}/\textcolor{blue}{64.6}                                         &\textcolor{blue}{85.6}/\textcolor{red}{65.0} \\
\textbf{NO}      &88.4/64.3            &\textcolor{blue}{93.1}/\textcolor{blue}{67.3} &\textcolor{red}{94.8}/\textcolor{red}{68.0}     &92.3/67.1     &{90.5}/{66.5}           &84.0/{63.0}                         &{88.1}/64.7 \\
\textbf{HO}       &70.3/46.5           &72.1/49.1      &73.8/50.7        &{74.5}/{52.1 }                      &{72.9}/{50.4}                      &\textcolor{red}{85.2}/\textcolor{red}{63.9}                        &\textcolor{blue}{83.4}/\textcolor{blue}{60.1} \\
\hline 
\textbf{ALL}     &80.0/55.4           &80.6/55.9      &82.7/57.9        &\textcolor{blue}{83.9}/{59.3}                &{78.6}/{56.2}                     &{83.0}/\textcolor{blue}{61.5}                 &\textcolor{red}{86.5}/\textcolor{red}{63.8} \\
\hline 
\end{tabular}
\label{tb::CA-2}
\end{table*}

\begin{figure*}[!htp]
    \centering
    \includegraphics[width=1.0\textwidth]{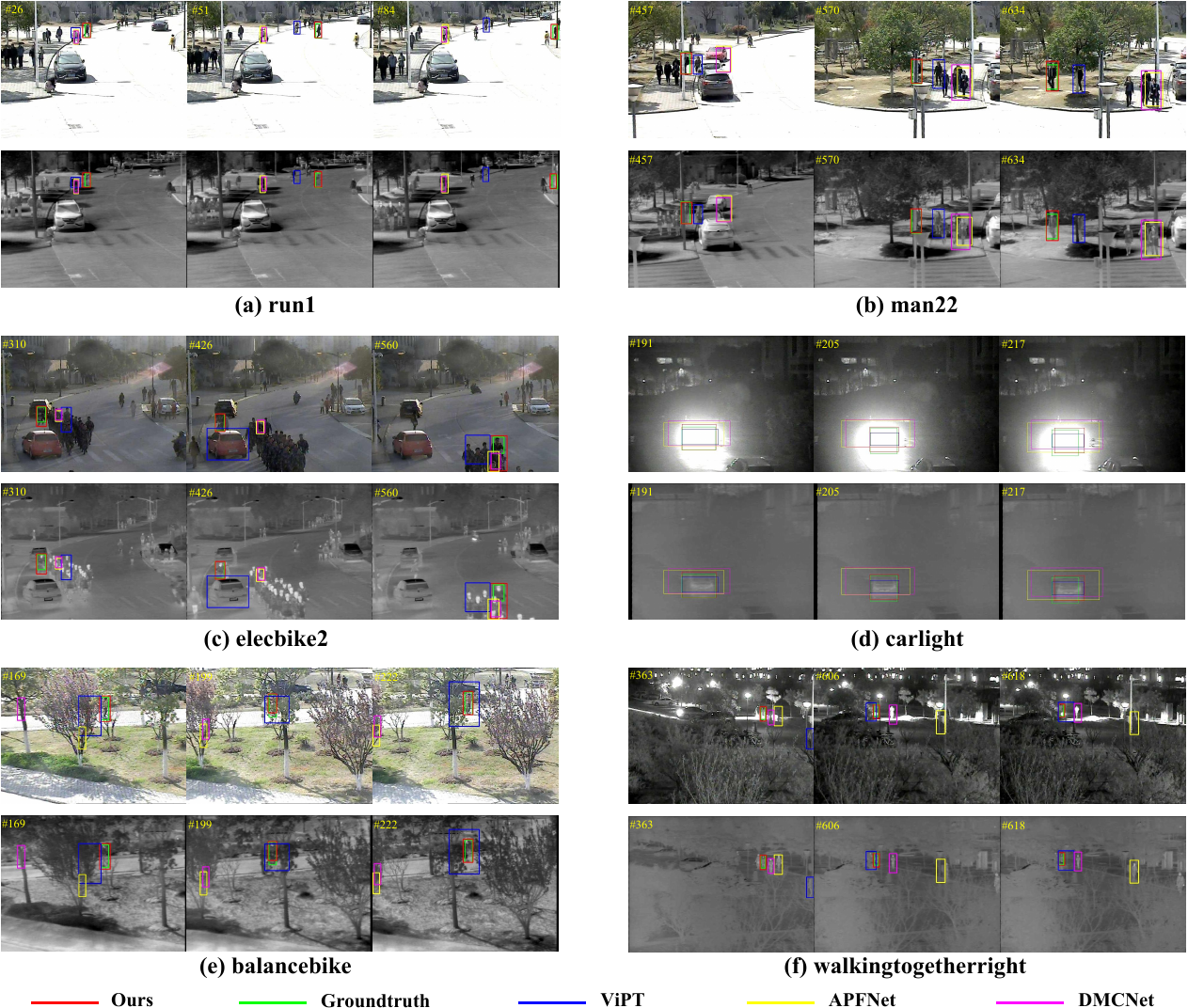} 
    \caption{Visual comparison of our tracker and three state-of-the-art trackers on six video sequences.}
    \label{fig::tracker_compare}
\end{figure*}

\subsubsection{Evaluation on RGBT234 dataset}
RGBT234 is the most widely used dataset in the field of RGBT tracking, and almost every RGBT method has been tested and evaluated on this dataset or its subsets. 
By observing Table~\ref{tb::result}, it is evident that our algorithm outperforms all the latest RGBT tracking algorithms. 
We compared our proposed method with ViPT\cite{zhu2023visual}, a transformer-based RGBT tracking algorithm recently published in CVPR 2023. 
Our proposed method improves the MPR/MSR scores up to +3.0\%, +2.1\%, superior performance over ViPT. 
As shown in Fig.~\ref{fig::metVSbase}, the existing method is adversely affected when the target undergoes significant changes in appearance compared to the original template. 
The part of search region similarity to the original appearance leads to scattered attention, resulting in inaccurate target localization. 
Our algorithm shows stronger robustness in extreme challenge scenarios compared to the baseline method and improves the MPR/MSR scores up to +3.5\%/+2.3\%.
This indicates that our method effectively addresses the limitations of existing approaches, specifically in terms of modal fusion and the lack of target variation information. 

Notably, the RGBT234 includes 12 challenge attributes: no occlusion (NO), partial occlusion (PO), heavy occlusion (HO), low illumination (LI), low resolution (LR), thermal crossover (TC), deformation (DEF), fast motion (FM), scale variation (SV), motion blur (MB), camera moving (CM) and background clutter (BC). To further verify the effectiveness of our method, we also compare the tracking performance of 12 challenge attributes, we show the results of our tracker against other state-of-the-art RGBT trackers in Table~\ref{tb::CA-2}, including MANET++~\cite{lu2021rgbt}, TFNet~\cite{zhu2021rgbt}, APFNet~\cite{xiao2022attribute}, DMCNet~\cite{lu2022duality}, DMSTM~\cite{zhang2023dual}, OSTrack~\cite{ye2022joint}-RGBT*. 
As shown in Table~\ref{tb::CA-2}, the results indicate that our method demonstrates excellent performance on the majority of challenges. 
Especially in the challenge of deformation (DEF) and low illumination (LI) conditions, our method exhibits a significant improvement compared to the second-ranked temporal-based approach, on DEF with an increase of +9.0\% in MPR and +5.8\% in MSR. This indicates that our temporal method can effectively maintain stable tracking of deforming targets when changes in appearance occur.
Furthermore, our method demonstrates significantly higher performance than the Baseline-RGBT* approach in the challenges of PO, TC, DEF, LI, CM, SV, and NO, which indicates a substantial improvement in the ability to model modal interactions and more accurately locate the target by leveraging appearance variation information. 

\subsubsection{Evaluation on LasHeR dataset}
Lasher is currently the largest and most challenging dataset for existing RGBT trackers. 
It comprises a substantial number of video sequences with various challenges, including similar appearances, partial occlusions and fast motion, etc. From Table~\ref{tb::result}, it is evident that our proposed method achieves impressive tracking results. 
Compared with recent temporal-based methods, it can be observed that while these approaches introduce temporal information, the lack of emphasis on the original target information hinders their performance from keeping up with state-of-the-art traditional methods.
In contrast, our proposed method improves the PR/SR scores up to +2.3\%, +1.2\% compared with the best traditional RGBT trackers. 
Compared to the baseline method with finetune, the PR/NPR/SR scores of our method are improved by +3.3\%/+3.2\%+2.6\%, and our method outperforms the latest transformer-based methods in terms of performance. 
The Lasher dataset also comprises numerous challenging data sequences, and it can be observed that existing temporal methods face limitations in their performance on this dataset. In contrast, our approach not only preserves the original reliable information but also introduces temporal information, leading to an enhanced performance on this dataset while maintaining the baseline performance.

\subsubsection{Compare the Latest and Most Competitive RGBT Trackers}
To further demonstrate the advanced performance of our algorithm, we compare it with recent methods as well as the top-performing RGB-T trackers. The comparison RGB-T trackers include DMCNet~\cite{lu2022duality}, DMSTM~\cite{zhang2023dual}, APFNet~\cite{xiao2022attribute}, ViPT~\cite{zhu2023visual}. DMCNet~\cite{lu2022duality} and APFNet~\cite{xiao2022attribute} represent the state-of-the-art performance of non-temporal methods on the RGBT234 dataset. DMSTM~\cite{zhang2023dual} represents the state-of-the-art performance of the temporal-based method on the RGBT234 dataset. ViPT~\cite{zhu2023visual} is the latest multimodal tracker that was published in CVPR 2023. 
As shown in Tab~\ref{tb::result}, our method achieves state-of-the-art performance on all three datasets. On the RGBT210 dataset, we outperform the DMCNet by +3.3\%/4.0\% in terms of MPR/MSR, respectively. On the RGBT234 dataset, compared with the second-best tracker we improve the MPR/MSR scores up to +2.6\%, +2.1\%, and on the LasHeR dataset we improve the PR/SR scores up to +2.3\%, +1.2\% which compared with ViPT. 
Compared to the second-ranked temporal-based method, we achieve excellent performance on three datasets. We improve the MPR/MSR scores up to +7.9\%, +5.7\% on the RGBT234, the PR/NPR/SR scores up to +11.7\%, +13.1\%, +13.7\% on the LasHeR. 

We also compare our tracker with three state-of-the-art trackers and visualize the tracking results under various challenging scenarios. 
As shown in Fig.~\ref{fig::tracker_compare}, our method demonstrates greater robustness compared to other trackers under multiple challenging scenarios. 
For example, the sequence in Fig.~\ref{fig::tracker_compare}(f) encompasses numerous challenges which include Low Resolution (LR), Partial Occlusion (PO), and Low Illumination (LI), our approach still tightly follows the target compared with other approaches. 
For another example, in some scenarios with strong illumination and severe occlusion, such as in sequences (d) and (e) shown in Fig.~\ref{fig::tracker_compare}, our tracker is still able to accurately track the target. 
In some cases, the ground truth labels may not accurately mark the tracked target. However, our method still accurately provides bounding boxes, as shown in sequences (b) and (c) in Fig.~\ref{fig::tracker_compare}. 

\subsection{Ablation Study}
To validate the effectiveness of major components in our method, we carry out the ablation study on the LasHeR and RGBT234.
\begin{table}[]\footnotesize
\renewcommand\arraystretch{1.5}
\caption{Ablation studies of our proposed module evaluate on LasHeR testingset. * indicates the tracker is re-trained.}
\centering
\resizebox{\linewidth}{!}{
\begin{tabular}{c|c|c|c}
\hline
\hline
\textbf{Method}    &\textbf{Precision}  &\textbf{NormPrec} &\textbf{Success} \\
\hline
Baseline-RGBT & 53.0 & 50.1 & 43.0 \\
Baseline-RGBT* & 64.1 & 60.2 & 51.1 \\
\hline
w/o Modal Mutual-Enhancement & 65.4 & 61.6 & 52.0 \\
w/o Dynamic token & 64.9 & 61.1 & 51.7 \\
\hline
\textbf{Full Model} & \textbf{67.4} & \textbf{63.4} &\textbf{53.7} \\
\hline
\end{tabular}}
\label{tb::as}
\end{table}

\subsubsection{Component Analysis}
As shown in Table~\ref{tb::as}, we conduct ablation studies on the LasHeR dataset to evaluate different designs of our module.

\textbf{Baseline-RGBT} denotes our extension of the original single-modality tracking baseline, where we replicate the data flow of its backbone network and concatenate the outputs of the backbone networks of both modalities. This concatenated feature is then fed into the tracking head, resulting in a multi-modality version of the tracker. 
We inherit the joint feature extraction and template matching method from the baseline network and employ a shared parameter backbone network for both modalities. 
The RGBT baseline network is evaluated by loading the pre-trained models provided by OSTrack and testing them on the Lasher testing dataset, which resulted in the obtained results. 
From the results, it can be observed that the transformer-based tracker achieves performance superior to almost all traditional network architectures for RGBT tracking by incorporating thermal infrared modality information and simple modal interactions. 

\textbf{Baseline-RGBT*} represents the same baseline network architecture, but with the difference that we trained the network using a large-scale multi-modal dataset-Lasher to make the network familiar with multi-modal data inputs. 
As a result, the evaluation results on RGBT datasets are expected to be higher compared to the original baseline. 
From the results, it is evident that specialized training for multi-modal tasks is indeed necessary. 
Training the network specifically for multi-modal tasks enables better adaptation to such tasks, ultimately leading to improved performance. Furthermore, we can also observe the potential of transformer-based trackers. Our modified multi-modal version tracker, after multimodal training, has nearly surpassed all existing methods based on CNN and those combining CNN and transformers.

\textbf{w/o Modal Mutual-Enhancement} denotes without performing cross-modal interaction at the hierarchical level and separately handling temporal information in each modality without modality-enhanced interaction. The performance improvement compared to our RGBT baseline version validates the importance of incorporating temporal information, even in the absence of cross-modal interaction at the hierarchical level. It demonstrates that the introduction of temporal information is crucial for enhancing tracking performance. 
Compared with the complete structure of STMT, we can also infer that the cross-modal interaction in our joint feature extraction and template matching stages helps to focus better on the target.

\textbf{w/o Dynamic token} denotes removing the temporal information with modality enhancement. Only in the joint feature extraction stage, do the RGB and TIR template tokens undergo modality-enhanced interaction. Compared to our RGBT baseline, we can observe that a small but reliable modality interaction can improve the effectiveness of multi-modal fusion. The performance drops compared with our full model version also indicate that the introduced modality-enhanced temporal information enhances the network's perception of target variations, benefiting the tracking process.
\begin{table}[]\footnotesize
\renewcommand\arraystretch{1.2}
\caption{Inserting layers of the proposed module test on RGBT234. We inserted our module in layers 4, 7, and 10, and applied dynamic token fusion at different stages of the module. The checkmark indicates the modules where we applied the temporal fusion component.}
\centering
\begin{tabular}{ccc|c|c}
\hline
\hline
\multicolumn{3}{c|}{Layer}  &\multirow{2}{*}{Precision}  &\multirow{2}{*}{Success} \\
4&7&10& & \\
\hline
           & &                          & 84.1 & 62.0 \\
\checkmark & &                          & 84.4 & 62.4 \\
           & \checkmark &               & 85.1 & 62.5 \\
           &            & \checkmark    & \textbf{86.5} & \textbf{63.8} \\
\checkmark & \checkmark &               & 84.7 & 62.8 \\
\checkmark & \checkmark & \checkmark    & 85.3 & 63.8 \\
\hline
\end{tabular}
\label{tb::as2}
\end{table}

\subsubsection{Inserting Layers of STMT Module}
The Dynamic token component has been applied at different inserting layers of our proposed module and summarizes the experimental results in Table~\ref{tb::as2}. By default, we retain modality enhancement in all three layers we applied it to. We can observe that inserting our module at different network layers already brings significant performance improvements which shows the importance of reliable modality enhancement in the joint feature extraction, particularly in the context of reliable template tokens. From Table~\ref{tb::as2}, we observe a significant performance improvement when applying temporal fusion in the deeper layers of the network. This indicates that temporal fusion is more effective in the deeper layers of features, as they are more conducive to capturing temporal dependencies. On the other hand, the shallower layers, with more noise in their features, may have a negative impact on the effectiveness of temporal fusion. Therefore, we adopt the configuration of three STMT modules as our final model but restrict the temporal fusion component to only take effect in the endmost module.

\subsection{Limitation Analysis}
Our STMT Transformer effectively addresses the issue of missing information caused by target appearance variations by introducing dynamic tokens, but there are certain limitations in its implementation approach. Our approach is achieved through a cross-attention mechanism. Although our design takes into account operating on a limited number of template-sized tokens, additional attention operations can still slow down the network's inference speed. The current dynamic tokens filtering strategy still uses a fixed threshold, which may not always result in optimal selection. To address the aforementioned limitations, future prospects can involve: 1) Replacing with a more efficient attention mechanism that reduces computational complexity without compromising performance. 2) Integrating attention mechanisms into the dynamic tokens filtering process to dynamically allocate attention weights to different dynamic tokens. This can help focus on more informative and relevant memories while reducing the influence of noisy or less useful ones. 

\section{Conclusion}
In this paper, we propose a novel Transformer RGBT tracking approach that enables an effective transformer network to focus on target appearance in RGBT tracking. Most previous RGBT methods either neglected the importance of temporal information or introduced temporal frames to fuse or replace initial templates, which can carry the risk of disrupting the original target appearance and accumulating errors over time. To alleviate these limitations, we propose a novel Transformer RGBT tracking approach, which mixes spatio-temporal multimodal tokens from the static multimodal templates and multimodal search regions in Transformer to handle target appearance changes, for robust RGBT tracking. In addition to introducing dynamic tokens into the training process, we have designed a temporal training strategy for training the temporal fusion component of the network. This strategy eliminates the need for additional network designs dedicated specifically to temporal training. Extensive experiments on three RGBT benchmark datasets show that the proposed approach maintains competitive performance compared to other state-of-the-art tracking algorithms. 


\bibliographystyle{IEEEtran}
\bibliography{reference.bib}

\end{document}